\documentclass{cta-author}

{}
{}
{}

\usepackage{amssymb}
\usepackage{latexsym}

\usepackage{url}

\begin{document}

\supertitle{IET Computer Vision SPECIAL ISSUE on
Visual Domain Adaptation and Generalization}

\title{ILGNet: Inception Modules with Connected Local and Global Features for Efficient Image Aesthetic Quality Classification using Domain Adaptation}

\author{\au{Xin Jin$^{1}$}, \au{Le Wu$^{1}$}, \au{Xiaodong Li$^{1}$}, \au{Xiaokun Zhang$^{1}$}, \au{Jingying Chi$^{2}$}, \au{Siwei Peng$^{2}$}, \au{\\Shiming Ge$^{3\corr}$}, \au{Geng Zhao$^{1}$}, \au{Shuying Li$^{4}$}}

\address{\add{1}{Department of Computer Science and Technology, Beijing Electronic Science and Technology Institute, Beijing 100070, China}
\add{2}{College of Information Science and Technology, Beijing University of Chemical Technology, Beijing 100029, China}
\add{3}{Institute of Information Engineering, Chinese Academy of Sciences, Beijing 100093, China}
\add{4}{The 16th Institute, China Aerospace Science and Technology Corporation, Xi'an 710100, China}
\email{geshiming@iie.ac.cn}\\
\url{https://github.com/BestiVictory/ILGnet}}

\begin{abstract}
In this paper, we address a challenging problem of aesthetic image classification, which is to label an input image as high or low aesthetic quality. We take both the local and global features of images into consideration. A novel deep convolutional neural network named ILGNet is proposed, which combines both the Inception modules and an connected layer of both Local and Global features. The ILGnet is based on GoogLeNet. Thus, it is easy to use a pre-trained GoogLeNet for large-scale image classification problem and fine tune our connected layers on an large scale database of aesthetic related images: AVA, i.e. \emph{domain adaptation}. The experiments reveal that our model achieves the state of the arts in AVA database. Both the training and testing speeds of our model are higher than those of the original GoogLeNet.
\end{abstract}


\maketitle

\section{Introduction}
\begin{figure}
\centering
\includegraphics[width=0.47\textwidth]{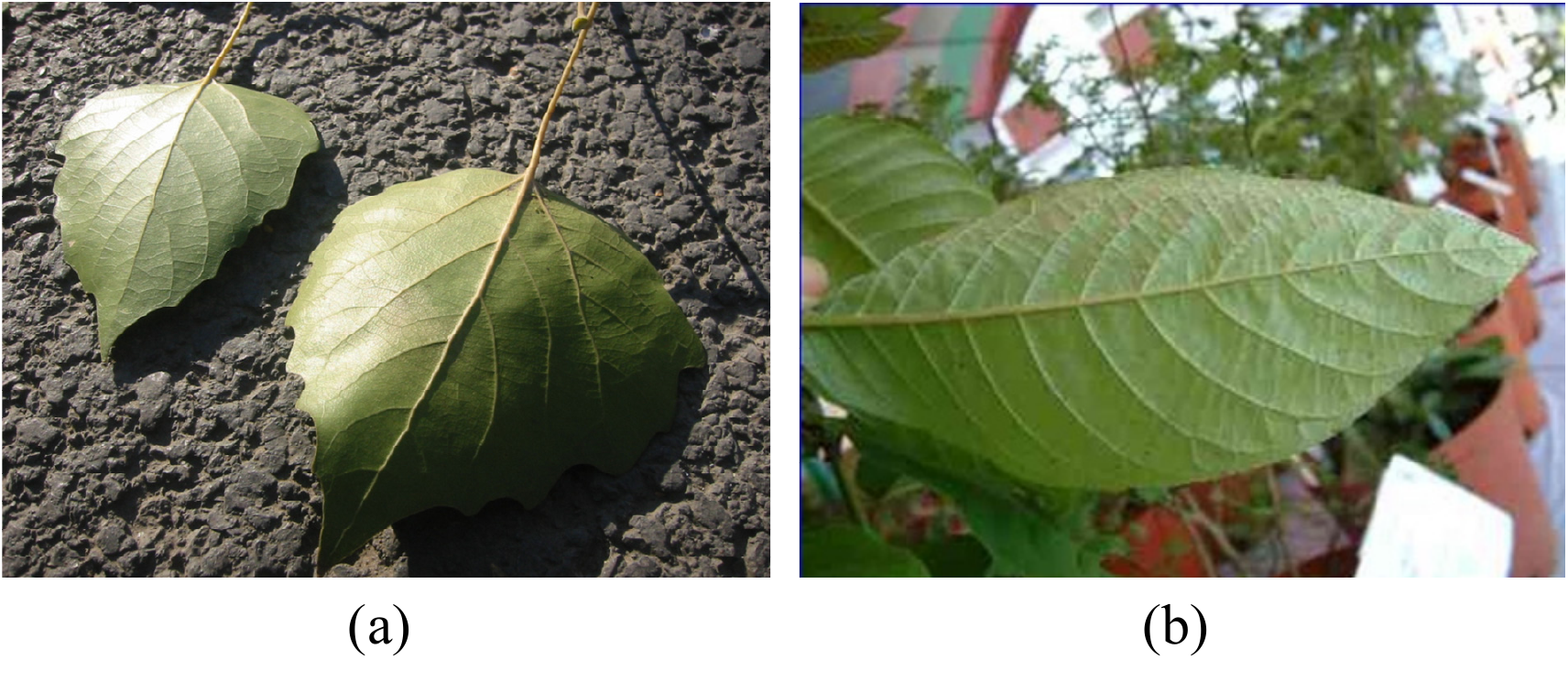}
\caption{The left image (a) is often considered as with higher aesthetic quality than the right one (b).}
\label{fig:example}
\end{figure}

Shooting good photos needs years of practice for photographers. However, it is often easy for people to classify an image into high or low aesthetic quality. As shown in Fig. \ref{fig:example}, the left image is often considered as with higher aesthetic quality than the right one.

Recently, smart phones, social networks and cloud computing boost the amount of images in the public or private cloud. People need a better way to manage their photos than ever before. A important ability of today's photo management software is to automatically recommend good photos from large amount of daily photos. Besides, aesthetic quality assessment can be used in the following scenarios:

\begin{enumerate}[(1)]
\item When you search images in the Internet, the aesthetic assessment engine can help to give you the ones with high aesthetic quality;

\item Nowadays, one may use their smart phones to shoot many photos everyday. Then, they struggle to find good photos from hundreds of photos so as to share selected ones in their social network such as Facebook, We chat, etc. In this scenario, aesthetic quality assessment can help them to make initial selection;

\item New image beautification software could be inspired by aesthetic quality assessment;

\item Large-scale on-line E-commerce platform need automatic designing of logo, banner or production introduction. The aesthetic quality classification can help to delete the ones with low aesthetic quality;

\item Automatic typesetting magazines, presentation documents, and scientific papers can rely on the aesthetic quality classification engines;

\item Other domains such as architecture, graphics, industry design, fashion design can use aesthetic quality assessment to classify hundreds of works into low or high quality.

\end{enumerate}

Today, image aesthetic quality classification is still a chanllenging problem. Typically, the following reasons make it challenging:

\begin{itemize}
\item Two classes of high and low aesthetic qualities contain large intra class differences;
\item Many high level aesthetic rules v.s. low level image features;
\item The subjective nature of human rating on aesthetic qualities of images.
\end{itemize}

Thus, people from computer vision, computational photography and computational aesthetics make this topic hot. In their early work, they design hand-crafted aesthetic image features, which are fed into a classification model or a regression model. Generic image features are also used in aesthetic quality classification. Today, deep convolutional neural networks are designed specially for aesthetic quality classification.

Recently, deep learning technologies have boosted the performance of many computer vision tasks \citep{WangTITS2017a}\citep{WangTITS2017b}\citep{LiTIP2018}\citep{JinAAAI2018}. Google proposed the inception module used in deep neural network architecture \citep{SzegedyCVPR2015}. The name of inception module are from the work of Lin et al \citep{LinCORR2013}. The inception module can be considered as a logical culmination of \citep{SzegedyCVPR2015}. It is inspired by Arora et al. \citep{AroraICML2014} in theory. In ILSVRC 2014, the architecture with inception module shows its benefits. The performance was significantly raised in the classification and detection challenges. 
However, in current literatures, inception modules has not been used in the aesthetic quality assessment to the best of our knowledge.

We propose to use inception modules for image aesthetics classification in this paper. A new deep convolutional neural network using Inception modules with connected Low and Global features is proposed, which is called ILGNet. Connecting intermediate layers directly to the output layers has show its value in recent work \citep{Maire2014} \citep{SzegedyCVPR2015}. In our ILGNet, the local features layers are connected to the global features layers. The ILGNet contains 13 layers with parameters and without counting pooling layers (4 layers). We use a pre-trained model on the ImageNet \citep{DengCVPR2009} as our initial model, which is trained for object classification of 1000 categories. Then, the inception modules are fixed and the connected local and global features layers are fine tuned on the AVA database, which is currently largest image aesthetics database \citep{MurrayCVPR2012}. We achieve the state of the art in the experiments on the AVA database \citep{MurrayCVPR2012}. Besides, the trained models and codes are available at github: \url{https://github.com/BestiVictory/ILGnet}.

The rest of this paper is organized as follows. In Section \ref{sec:RelatedWork}, we review the related work. In Section \ref{sec:Method}, we describe our proposed ILGNet in details. Then the experimental settings, results and comparisons with state-of-the-art methods are presented in Section \ref{sec:Experiments}. Finally, we give a conclusion in Section \ref{sec:Conclusion}.

\begin{figure*}[!htb]
\begin{center}
\includegraphics[width=1 \textwidth]{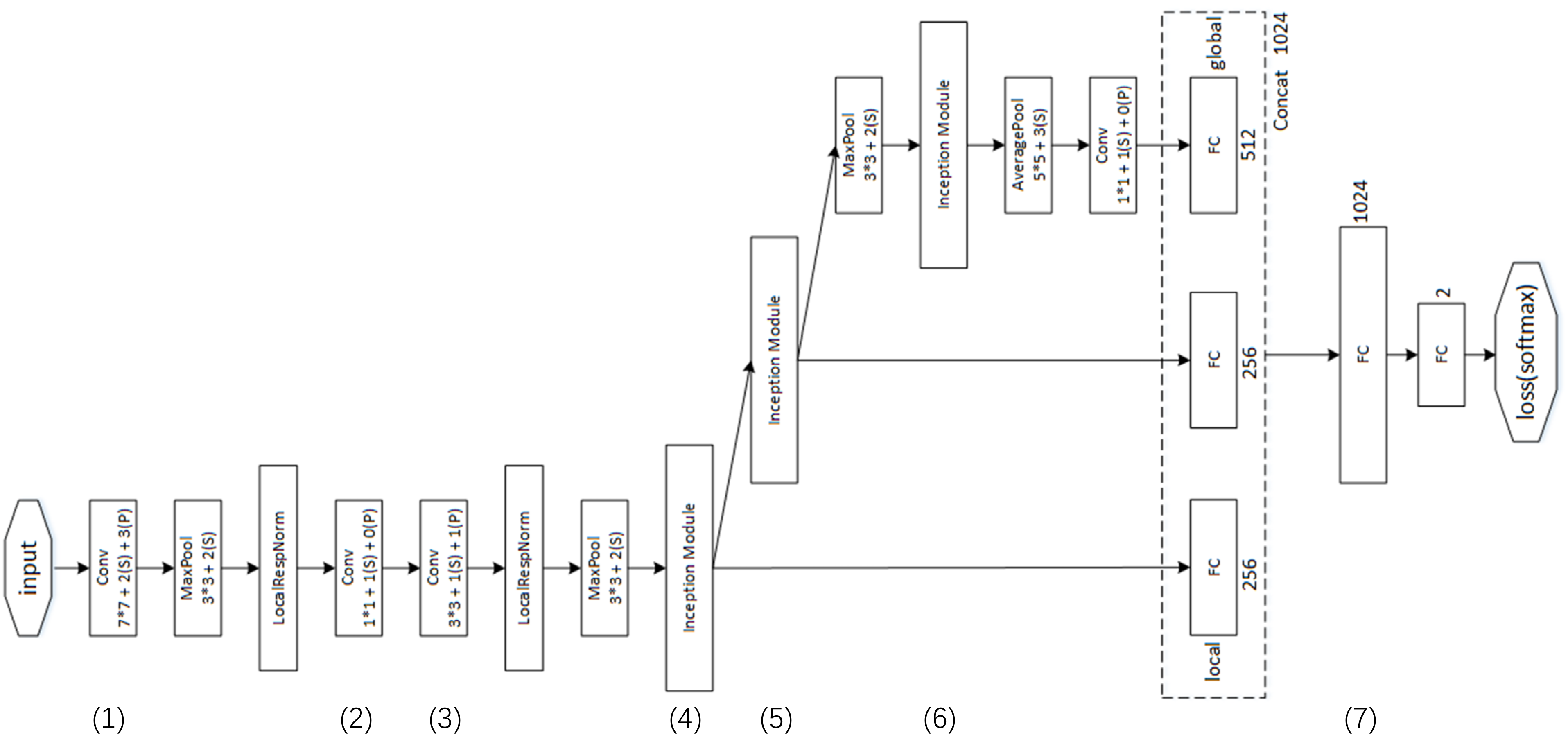}
\end{center}
\caption{The ILGNet architecture: Inception with connected Local and Global layers. We build this network on the first 1/3 part of GoogLeNetV1 \citep{SzegedyCVPR2015} and batch normalization, which is a important feature of GoogLeNetV2 \citep{IoffeICML2015}. 1 pre-treatment layer and 3 inception modules are used. We use the first 2 inception modules to compute the local features and the last one to compute global features.
Connecting intermediate layers directly to the output layers has show its value in recent work \citep{Maire2014} \citep{SzegedyCVPR2015}. Thus, we build a concat full connected layer of 1024 dimension which connect 2 layers of local features and a layer of global features. The output layer indicate the probability of low or high aesthetic quality. The ILGNet contains 13 layers with parameters and without counting pooling layers (4 layers). In Section \ref{sec:Experiments}, we use the labels (1)-(7) to demonstrate the visualization results.}
\label{fig:ILGNet}
\end{figure*}

\section{Previous Work} \label{sec:RelatedWork}
The related work of our task can be categorised into the traditional image quality assessment, the subjective image aesthetic quality assessment using hand-crafted features and deep learning.

\subsection{Traditional Image Quality Assessment}
Traditional image quality assessment is to assess the objective image quality, which may be distorted or influenced during the imaging, compression and transmission. Distortions such as ringing, blur, ghosting, smearing, blocking, mosaic, jerkiness are measured \citep{MaiCVPR2016}. The human perception of aesthetics can not be well modeled by these low-level features and metrics.

\subsection{Hand-crafted Features for Subjective Image Aesthetic Quality Assessment}
Subjective image aesthetic quality assessment is to automatically distinguish an image to low or high aesthetic quality. Some of them can give a numerical assessment. They often contains the three steps in the following:

\begin{itemize}

\item A database of images is collected. Then they often manually label each with two label: \emph{good} for images with high aesthetic quality, and \emph{low} for images with low aesthetic quality. Some make psychological experiments so as to get numerical assessment for part of the images in the database. 

\item Image features for aesthetic quality assessment are designed such as simplicity, visual balance and rule of third \citep{DattaECCV2006}\citep{KeCVPR2006}\citep{LuoECCV2008}
\citep{LiJSTSP2009}
\citep{BhattacharyaMM2010}
\citep{JiangICME2010}\citep{LiICIP10}
\citep{JinECCV2010}
\citep{GrayECCV2010}
\citep{ChenTIP2015}\citep{DharCVPR2011}
\citep{JoshiSPM2011}\citep{NishiyamaCVPR2011}\citep{LuoICCV2011}\citep{TangTMM2013}
\citep{WuICCV2011}
\citep{KhanCA12}\citep{NiuTCSVT2012}.
Generic image features which are previously used for object recognition are also used for aesthetic quality assessment, such as low level image features\citep{TongPCM2004}, bag of visual words \citep{SuMM2011}\citep{SuTMM2012}, and Fisher Vector \citep{MarchesottiICCV2011}.

\item Machine learning technologies such as random forest, support vector machine and boosting are used for image aesthetic quality assessment. They use the aesthetic database to train a classifier so as to classify an image into low or high aesthetic quality. They regress the human rating score to give a numerical assessment of the aesthetic quality of an image.

\end{itemize}

\subsection{Subjective Image Aesthetic Quality Assessment using Deep Learning}

Recently, deep learning technologies have boosted the performance of many computer vision tasks \citep{WangTITS2017a}\citep{WangTITS2017b}\citep{LiTIP2018}\citep{JinAAAI2018}\citep{LinGRSL2017}.
Deep belief network and deep convolutional neural network have been used for image aesthetics assessment. The performance has been significantly improved compared with traditional methods.
\citep{KarayevBMVC2014}\citep{LuMM2014}
\citep{LuICCV2015}\citep{LuTMM2015}
\citep{DongNC2015}
\citep{WangSP2016}\citep{MaiCVPR2016}
\citep{KongECCV2016}
\citep{KaoTIP2017}
\citep{WangIJCNN2017}
\citep{MaCVPR2017}.  

Most of the above work us the AlexNet architecture \citep{KrizhevskyNIPS2012}, which contains 8 layers with 5 convolutional layers and 3 full-connected layers or VGG \citep{Simonyan14c}. Inspried by the good performance of GoogLeNet in the ImageNet, which argues that deeper architectures enable to capture large receptive field. We can extract local image features and the global features of the image layout. Connecting intermediate layers directly to the output layers has show its value in recent work \citep{Maire2014} \citep{SzegedyCVPR2015}. Both the local features and the global features can be extracted by inception modules. Thus, we change the GoogLeNet by connecting the intermediate local feature layers to the global feature layer.


\section{ILGNet for Image Aesthetic Quality Classification} \label{sec:Method}
The details of the proposed ILGNet are described in this section. The ILGNet contains 13 layers with parameters and without counting pooling layers (4 layers). The network contains one pre-treatment layer and 3 inception modules. Two intermediate layers of local features are connected to a layer of global features, which makes a 1024 dimension concat layer. The output layer indicate the probability of low or high aesthetic quality. The basic ILGNet is built on the first 1/3 part of of GoogLeNetV1 \citep{SzegedyCVPR2015} and batch normalization, which is a important feature of GoogLeNetV2 \citep{IoffeICML2015}.

\subsection{The Inception Module}


The InceptionV1 module is proposed by GoogLeNetV1 \cite{SzegedyCVPR2015}.  The main ideas of the Inception module are:

\begin{enumerate}

\item Convolution kernels with different sizes represent receptive fields with difference sizes. This design means fusing features of different scales.

\item The kernel sizes are set to $1*1$, $3*3$ and $5*5$ so as to align the features conveniently. The stride is 1. The pad is set to 0, 1 ,2. 

\item The features extracted by the higher layer are increasingly abstract. The receptive field involved by  each feature is larger. Thus, the ratio of $3*3$ and $5*5$ kernels should be increased.

\end{enumerate}

After InceptionV1, Google proposed InceptionV2 and InceptionV3, which adopt factorization of convolutions and improved normalization. Then, InceptionV4 considered the residue network, which surpassed its ancestor GoogLeNet on the ImageNet benchmark.

%


\subsection{Image Aesthetic Quality Classification}
The convolution layers inside ILGNet us rectified linear activation. The size of the input receptive field of ILGNet is $224 \times 224$ in color images with zero mean \citep{SzegedyCVPR2015}. We use the first 2 inception modules to compute the local features and the last one to compute global features with 2 max pooling and 1 average pooling. Then, we build a concat full connected layer of 1024 dimension which connect 2 layers of local features (each layer is 256 dimension) and a layer of global features (512 dimension). The output layer is bypass a softmax layer to indicate the probability of low or high aesthetic quality. 

The ILGnet is based on GoogLeNet. Thus, it is easy to use a pre-trained GoogLeNet for large-scale image classification problem and fine tune our connected layers on an large scale database of aesthetic related images: AVA \citep{MurrayCVPR2012}, i.e. \emph{domain adaptation}.

%

\section{Experimental Results} \label{sec:Experiments}
We test the effectiveness of our ILGNet in the public AVA datebase \citep{MurrayCVPR2012}, which is specially designed for aesthetics analysis. The comparison experiments with the state of the art methods on aesthetic quality classification are shown in this section. Most of them use deep convolutional neural networks. The main training parameters of the Caffe package \cite{JiaArXiv2014} are listed in Table \ref{tb:parameter}.

\subsection{Database and Comparison Protocols}
%
The Aesthetic Visual Analysis database \citep{MurrayCVPR2012} is a list of image ids from DPChallenge.com, which is a on-line photography social network. There are total 255,529 photos, each of which is rated by 78-549 persons, with an average of 210. The range of the scores rated by human is 1-10. We use the same protocols to those of previous work. They often use two sub database of AVA.

\begin{itemize}

\item AVA1: The score of 5 are chosen as the threshold to distinguish the AVA to high (good) and low (bad) aesthetics quality. 74,673 images are labelled as bad photos. 180,856 are labelled as good photos. We randomly split the AVA database into training set (234,599) and testing set (19,930) \citep{MurrayCVPR2012}\citep{WangSP2016}
\citep{WangIJCNN2017}\citep{KongECCV2016}\citep{LuTMM2015}\citep{LuICCV2015}\citep{MaiCVPR2016}.

\item AVA2: The images in the AVA database are sorted according to their mean scores of the aesthetic quality. Then the top 10\% images are labelled as good. The bottom 10\% are labelled as bad. Thus, there are totally 51,106 images from AVA database. The 51,106 images are randomly divided into 2 sets with equal numbers, which are the training set and testing set respectively \citep{LuoECCV2008}\citep{LoICPR2012}\citep{DattaECCV2006}\citep{KeCVPR2006}\citep{MarchesottiICCV2011}\citep{DongNC2015}\citep{DongMMM2015}\citep{WangSP2016}.
\end{itemize}



%
%
	
\begin{figure}
\centering
\includegraphics[height=14.5cm]{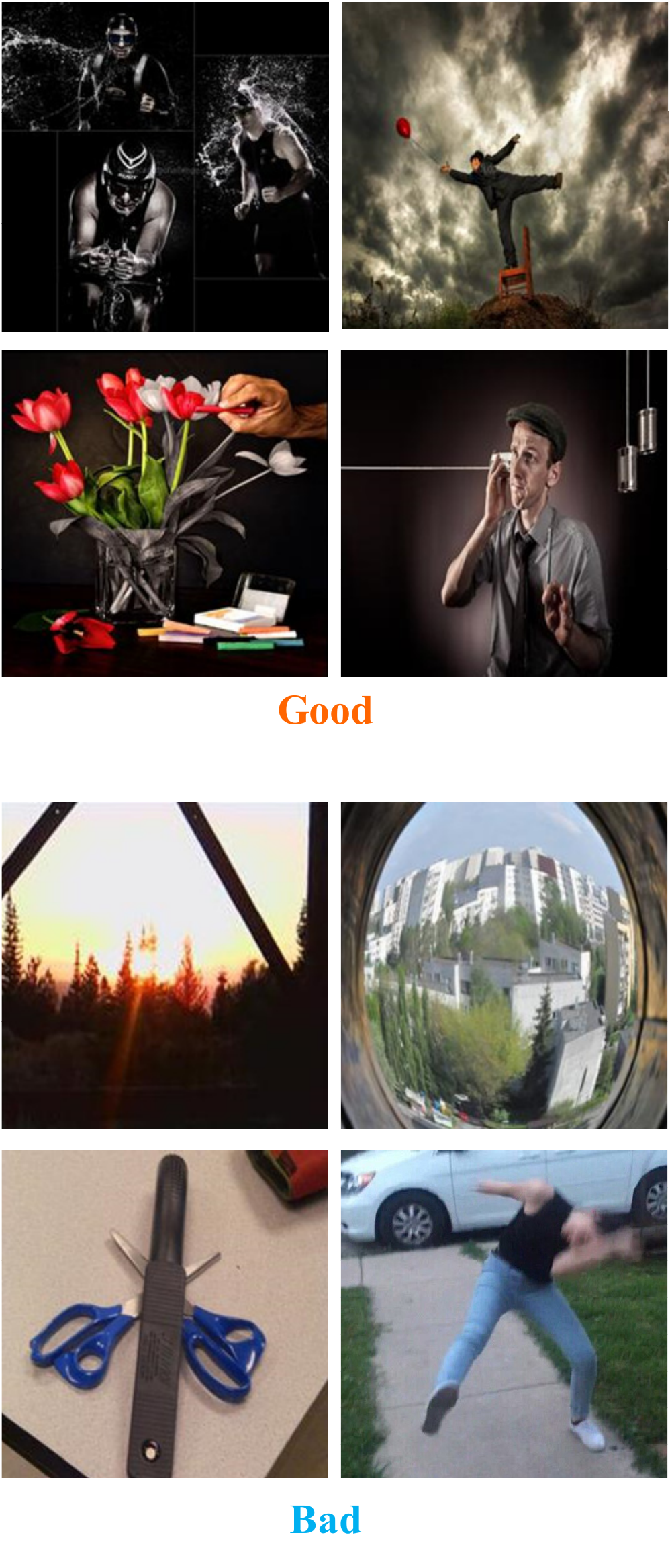}
\caption{We use the trained ILGNet to label images with good or bad, which indicates high or low aesthetic quality, respectively.}
\label{fig:Classification}
\end{figure}

\subsection{Classification Results}

As shown in Fig. \ref{fig:Classification}, We use the trained ILGNet to label images with good or bad, which indicates high or low aesthetic quality, respectively. Differences between low-aesthetic images and high-aesthetic images heavily lie in the amount of textures and complexity of the entire image \citep{LuTMM2015}.

\begin{table}[!t] \footnotesize
\renewcommand{\arraystretch}{1.3}
\caption{The main training parameters of the Caffe package.}
\label{tb:parameter}
\centering
\begin{tabular}{c||c||c||c}
\hline
Parameters & AVA1 ($\delta = 0$)  & AVA1 ($\delta = 1$) & AVA2 \\
\hline\hline
base\_ lr 	& 0.0001 & 0.00001 & 0.00001\\
\hline
lr\_ policy & "step" & "step" & "step" \\
\hline
stepsize 	& 100000 & 19000 & 13325\\
\hline
gamma 		& 0.96   & 0.96 & 0.96\\
\hline
max\_ iter	& 475000 & 760000 & 533000\\
\hline
momentum	& 0.9	 & 0.9 & 0.9\\
\hline
weight\_ decay & 0.0002 & 0.0002 & 0.0002\\
\hline
\end{tabular}
\end{table}

\begin{table}[!t]
\renewcommand{\arraystretch}{1.3}
\caption{The Classification Accuracy in AVA1 database.}
\label{tb:AVA1}
\centering
\begin{tabular}{c||c||c}
\hline
\bfseries Methods & \bfseries  $\delta = 0$ &  $\delta = 1$\\
\hline\hline
Traditional method\citep{MurrayCVPR2012} & 66.70\% & 67.00\%\\
\hline
RAPID \citep{LuMM2014}  & 69.91\% & 71.26\%\\
\hline
RAPID-E \citep{LuTMM2015}  & 74.46\% & 73.70\%\\
\hline
Multi-patch \citep{LuICCV2015} & 75.41\% & --\\
\hline
AROD \citep{SchwarzArXiv2016} & 75.83\% & --\\
\hline
Multi-scene \citep{WangSP2016} & 76.94\% & --\\
\hline
Comp.-prev. \citep{MaiCVPR2016} & 77.10\% & 76.10\%\\
\hline
AADB \citep{KongECCV2016} & 77.33\% & --\\
\hline
BDN \citep{WangIJCNN2017} & 78.08\% & 77.27\%\\
\hline
Semantic-based \citep{KaoTIP2017} & 79.08\% & 76.04\%\\
\hline
A-Lamp \citep{MaCVPR2017} & 82.5\% & --\\
\hline
\textbf{ILGNet-without-Inc.}  & \textbf{75.29\%} & \textbf{73.25\%}\\
\hline
\textbf{1/3 GoogLeNetV1-BN}  & \textbf{80.74\%} & \textbf{79.09\%}\\
\hline
\textbf{ILGNet-Inc.V1-BN}  & \textbf{81.68\%} & \textbf{80.71\%}\\
\hline
\textbf{ILGNet-Inc.V3}  & \textbf{81.71\%} & \textbf{80.65\%}\\
\hline
\textbf{ILGNet-Inc.V4}  & \textbf{82.66\%} & \textbf{80.83\%}\\
\hline
\end{tabular}
\end{table}


\begin{table}[!t]
\renewcommand{\arraystretch}{1.3}
\caption{The Classification Accuracy in AVA2 database.}
\label{tb:AVA2}
\centering
\begin{tabular}{c||c}
\hline
\bfseries Methods & \bfseries Accuracy\\
\hline\hline
Subject-based \cite{LuoECCV2008} & 61.49\%\\
\hline
EfficientAssess \cite{LoICPR2012} & 68.13\%\\
\hline
Generic-based \cite{MarchesottiICCV2011} & 68.55\%\\
\hline
Compt.-based \cite{DattaECCV2006} & 68.67\%\\
\hline
High-level \cite{KeCVPR2006}  & 71.06\%\\
\hline
Multi-level \cite{DongNC2015} &78.92\%\\
\hline
Query-dependent \cite{TianTMM2015} & 80.38\%\\
\hline
DCNN-Aesth-SP \cite{DongMMM2015} &83.52\%\\
\hline
Multi-scene \cite{WangSP2016} &84.88\%\\
\hline
\textbf{ILGNet-without-Inc.}  & \textbf{79.64\%}\\
\hline
\textbf{1/3 GoogLeNetV1-BN}  &  \textbf{82.26\%}\\
\hline
\textbf{ILGNet-Inc.V1-BN} & \textbf{85.50}\%\\
\hline
\textbf{ILGNet-Inc.V3}  & \textbf{85.51}\%\\
\hline
\textbf{ILGNet-Inc.V4}  & \textbf{85.53}\%\\
\hline
\end{tabular}
\end{table}

\begin{table}[!t] \footnotesize
\renewcommand{\arraystretch}{1.3}
\caption{The Efficiency Comparison in AVA1 database.}
\label{tb:time}
\centering
\begin{tabular}{c||c||c||c}
\hline
\bfseries Methods & \bfseries  Accuracy $\delta = 0$ &  Training Time & Test Time\\
\hline\hline
Full GoogLeNetV1-BN & \textbf{82.36\%} & 16 days & 0.84s\\
\hline
2/3 GoogLeNetV1-BN & 81.72\% & 11 days & 0.57s\\
\hline
1/3 GoogLeNetV1-BN & 80.74\% & \textbf{4 days} & 0.33s\\
\hline
\textbf{ILGNet-Inc.V1-BN}  & 81.68\% & \textbf{4 days} & \textbf{0.31s}\\
\hline
\end{tabular}
\end{table}

\begin{figure}
\centering
\includegraphics[height=5.3cm]{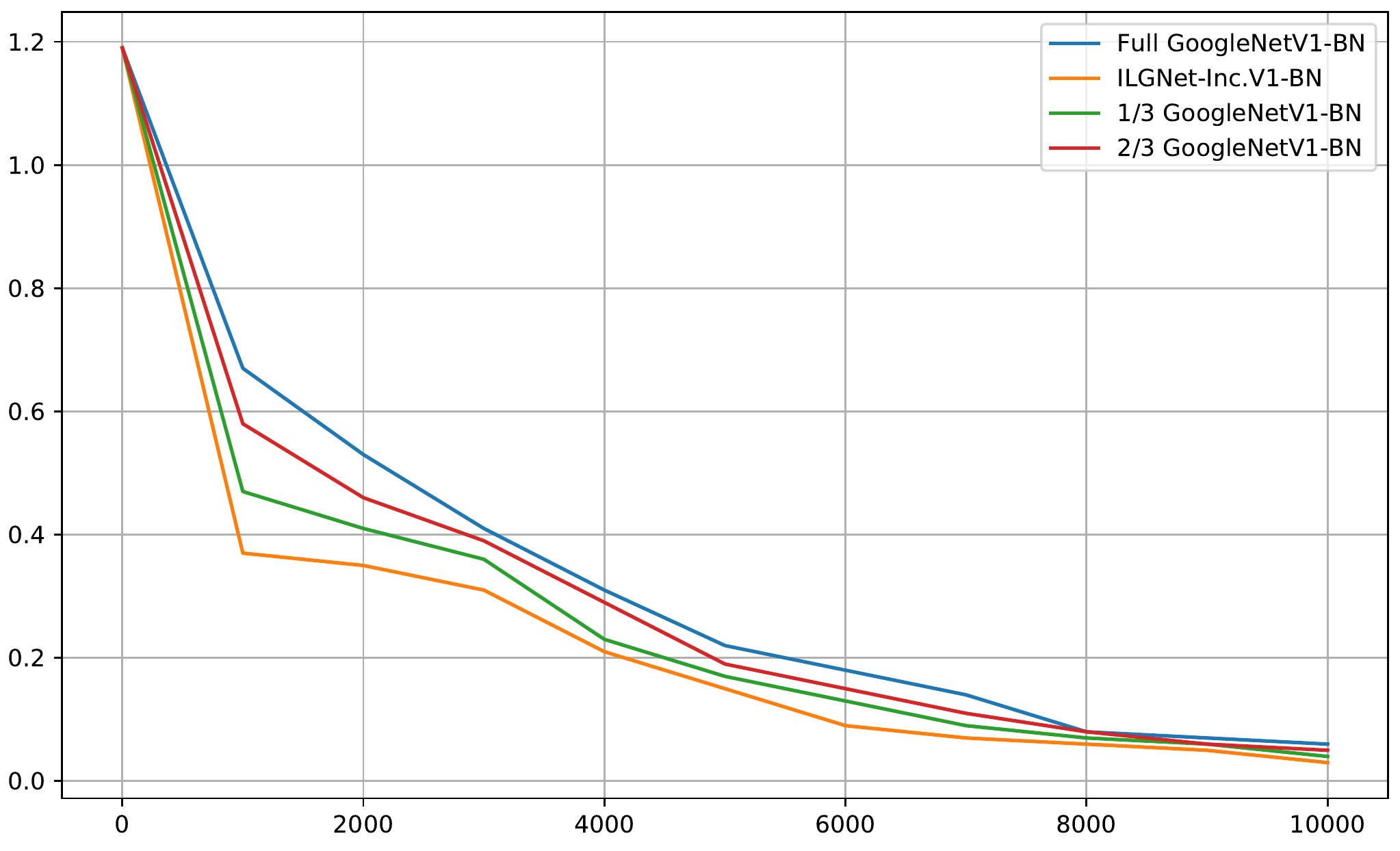}
\caption{Loss vs. epoch of our ILGNet-Inc.V1-BN,  1/3, 2/3 and full GoogLeNetV1-BN. in AVA1 database.}
\label{fig:epoch}
\end{figure}

\begin{figure*}
\centering
\includegraphics[height=20cm]{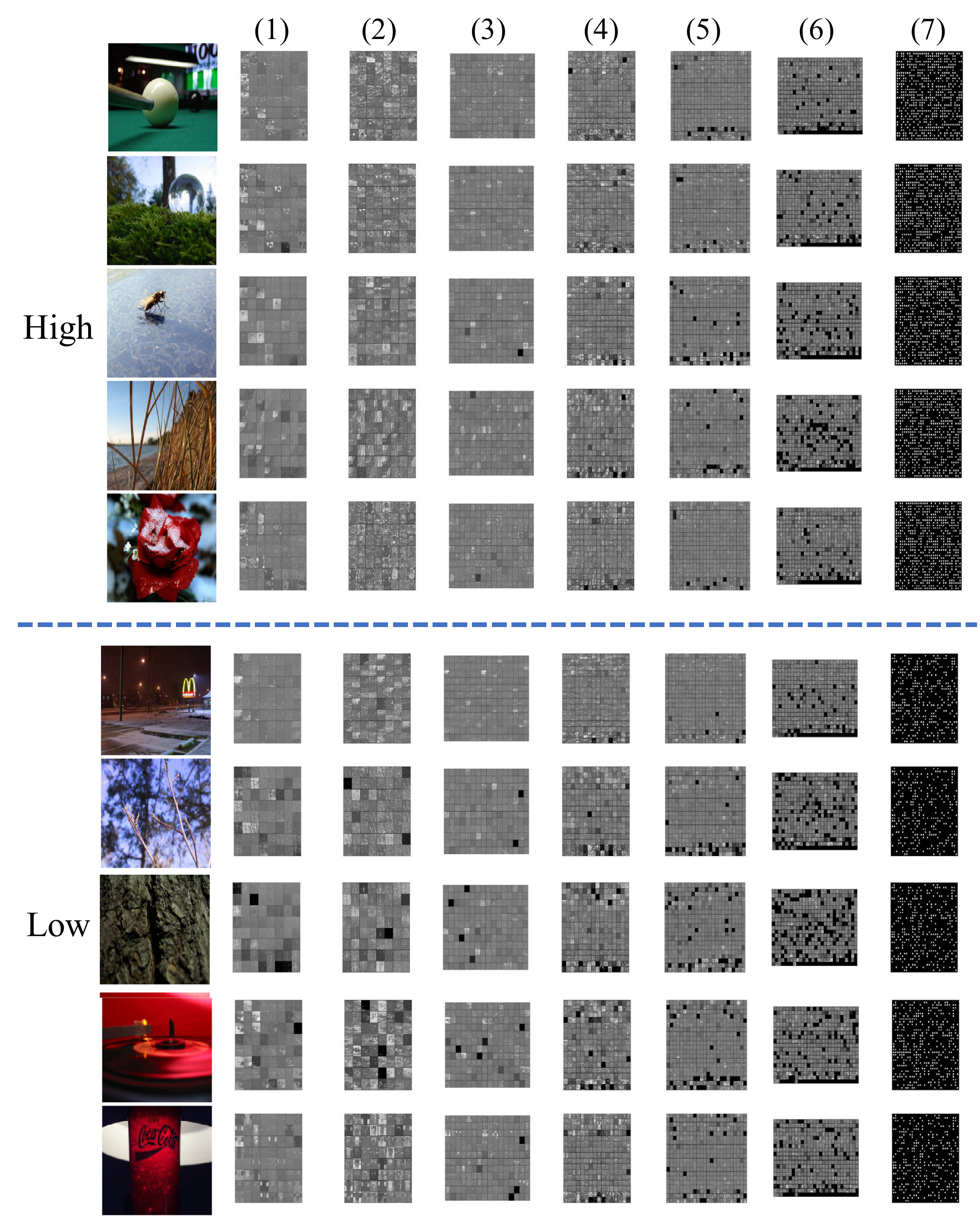}
\caption{The extracted features using the ILGNet-Inc.V1-BN of good and bad photos. The labels of (1)-(7) means the same in Fig. \ref{fig:ILGNet}. We have an interesting observation that in the last layer, the density of the active features are often higher in the ones with high aesthetic quality than those with low aesthetic quality.}
\label{fig:FeatureVis}
\end{figure*}

%

The original ILGNet is build on the first 1/3 of GoogLeNet V1, as shown in Fig. \ref{fig:ILGNet}. We add batch normalization (GoogLeNet V2 \citep{SzegedyCVPR2015} features), which form our ILGNet-Inc.V1-BN. After that we further build our ILGNet on the first 1/3 of recent GoogLeNet V3 \citep{SzegedyCVPR2016} and V4 \citep{SzegedyAAAI2017}, which form our ILGNet-Inc.V3 and ILGNet-Inc.V4. The test results in the AVA1 database are shown in Table \ref{tb:AVA1}. Our ILGNet-Inc.V4 outperforms the other DCNN based methods and achieve the state of the art accuracy: 82.66\%.

The above is the case of $\delta=1$. Similar results are shown when $\delta=1$. In the original test protocol \citep{MurrayCVPR2012}, they set $\delta=1$ in the training set, there are 7,500 low-quality images and 45,000 high-quality images. For the testing images, they fix $\delta$ to 0, regardless what $\delta$ is used for training. We have tested five network architectures on $\delta=0$ and $\delta=1$. The results are shown in Table \ref{tb:AVA1}. The ambiguity image samples are removed from the training set. Ambiguity images are still in the test set. Thus, the decreasing of accuracy is reasonable. We still achieve the state of the arts performance when delta =1.

To verify the effectiveness of inception module, we test a modified network of ILGNet-Inc.V1-BN: the ILGNet-without-Inc., in which we replace all the inception module with corresponding ordinal convolutional layer that is adaptive with the original pre and next layers. The performance (75.29\%) of this INGNet-without-Inc. is significantly worse than that (81.68\%) of the ILGNet-Inc.V1-BN. This verifies the usefulness of the inception module in capture features of both local patch and global view.

To verify the effectiveness of the connected local and global layer, we compare our ILGNet-Inc.V1-BN with the first 1/3 of original GoogLeNet with batch normalization: 1/3 GoogleNetV1-BN. The performance (80.74\%) of the 1/3 GoogleNetV1-BN on AVA1 is also worse than that (81.68\%) of the ILGNet-Inc.V1-BN. This verifies the usefulness of our proposed connected local and global layer.

The test results in the AVA2 database are shown in Table \ref{tb:AVA2}. Our ILGNet-Inc.V4 outperforms the other DCNN based methods and achieve the state of the art accuracy: 85.53\%.


\subsection{The Efficiency Comparison}
We take the ILGNet-Inc.V1-BN as an example to compare the efficiency with the first 1/3, 2/3 and full GoogLeNetV1 plus batch normalization. The time costs are summarized in Table \ref{tb:time}.  The test time is the average time on the test set of AVA1 and a Nvidia GTX980ti card. The time cost of both training and test of the ILGNet-Inc.V1-BN are significantly less than those of full GoogLeNetV1-BN with only a little reduction of the classification accuracy. This makes the aesthetic assessment model more easily to be integrated into mobile and embedded systems. 

The performance of our ILGNet-Inc.V1-BN is better than that of 1/3 GoogLeNetV1-BN. The training and test times of our ILGNet-Inc.V1-BN is similar as those of 1/3 GoogLeNetV1-BN. This is because that our ILGNet-Inc.V1-BN is built on the 1/3 GoogLeNetV1-BN, which has similar computational efficiency as ours. With our strategy of connected local and global layer, our ILGNet-Inc.V1-BN can even achieve nearly the same performance (81.68\%) to that (81.72\%) of 2/3 GoogleNetV1-BN. While the training and test times of 2/3 GoogleNetV1-BN are much more than those of our ILGNet-Inc.V1-BN.  In addition, we show the loss vs. epoch curves in Fig. \ref{fig:epoch}. Our ILGNet-Inc.V1-BN achieve the fastest convergence speed, which further verifies the efficiency of our method.

\subsection{The Features Visualization}

%

The extracted features using the ILGNet-Inc.V1-BN are visualized in Fig. \ref{fig:FeatureVis} for images with high and low aesthetic quality. The proposed ILGNet-Inc.V1-BN can be used to compute the low level features and high level features. The connected layer of local and global features are shown at last. It can be observed that the last feature maps are nearly binary pattens. We have an interesting observation that in the last layer, the density of the active features are often higher in the ones with high aesthetic quality than those with low aesthetic quality. This verifies that the extracted features can well represent the aesthetic quality.

\section{Conclusion and Discussion} \label{sec:Conclusion}
We propose a new DCNN called ILGNet for subjective image aesthetic quality classification. The ILGNet is derived from part of GoogLeNet. Thus, it can be used for domain adaptation from image classification to image aesthetic quality classification. The bottom features are shared for this two tasks. The high level features together with 2 inception modules are fine tuned for aesthetic quality classification. We fixed the shared inception layers of a pre-trained GoogLeNet model on the ImageNet \citep{DengCVPR2009} and fine tune the connected layer on the AVA database \citep{MurrayCVPR2012}. The proposed ILGNet outperforms the state of the art methods in AVA database. 

In the future work, we will address the following problems.

\textbf{Hyperparameter}. We hope that some architecture parameters such as the number of layers and the number of nodes on the full connected layers can also be automatically determined from the training on large-scale aesthetic database. 

\textbf{Composition}. Now the input image is scaled to a fixed size of 224*224, which loses high quality local image patches and destroys the composition aesthetics of the original image. In the future work, we will use technologies such as spatial pyramid pooling to handle this limitation. 

\textbf{Database Bias}. Because of the bias of AVA database (the number of high quality images is higher than that of low quality images), we will explore other aesthetic criteria such as numerical assessment or ranking in the future.

\textbf{Consensus}. The aesthetic assessment is a subjective task in nature. The mean score of an image can describe the overall impression to some extent with consensus. In the future work, we need to assess the aesthetic quality from more views such as score distribution, photography attributes, aesthetic caption.

\section*{Acknowledgments}
This work is partially supported by the National Natural Science Foundation of China (Grant Nos. 61402021, 61401228, 61402463, 61772513), the Science and Technology Project of the State Archives Administrator (Grant No. 2015-B-10), the open funding project of State Key Laboratory of Virtual Reality Technology and Systems, Beihang University (Grant No. BUAA-VR-16KF-09), the Fundamental Research Funds for the Central Universities (Grant No. 3122014C017), the China Postdoctoral Science Foundation (Grant No. 2015M581841), and the Postdoctoral Science Foundation of Jiangsu Province (Grant No. 1501019A).

Parts of this paper have previously appeared in our previous work \cite{JinWCSP2016}. This is the extended journal version of that conference paper.  The main differences between this archival version and the conference version are:

\begin{enumerate}
\item The title has been changed to capture essential idea of our work.

\item A clearer explanation of the power of the Inception module.

\item The performance of our ILGNet is increased to the state of the art (79.25\% to 82.66\% with the new network combination ILGNet-Inc.V4 reported in this version).

\item More experimental results are shown and more related methods (including the state of the art method published after the publication of our conference paper) are compared.

\item We have published our newly trained models and codes at \url{https://github.com/BestiVictory/ILGnet}
\end{enumerate}

\bibliographystyle{IEEEbib}
\bibliography{refs}

\end{document}